# Optimization of Dynamic Mobile Robot Path Planning based on Evolutionary Methods

Masoud Fetanat
Department of Electrical Engineering
Sharif University of Technology
Tehran, Iran
fetanat@ee.sharif.edu

Sajjad Haghzad
Department of Electrical Engineering
Sharif University of Technology
Tehran, Iran
sajjad_haghzad@ee.sharif.edu

Saeed Bagheri shouraki
Department of Electrical Engineering
Sharif University of Technology
Tehran, Iran
bagheri_s@sharif.edu

*Abstract* - This paper presents evolutionary methods for optimization in dynamic mobile robot path planning. In dynamic mobile path planning, the goal is to find an optimal feasible path from starting point to target point with various obstacles, as well as smoothness and safety in the proposed path. Pattern search (PS) algorithm, Genetic Algorithm (GA) and Particle Swarm Optimization (PSO) are used to find an optimal path for mobile robots to reach to target point with obstacle avoidance. For showing the success of the proposed method, first they are applied to two different paths with a dynamic environment in obstacles. The first results show that the PSO algorithms are converged and minimize the objective function better that the others, while PS has the lower time compared to other algorithms in the initial and modified environment. The second test path is in the z-type environment that we compare the mentioned algorithms on it. Also in this environment, the same result is repeated.

## I. INTRODUCTION

Many researchers have interests on autonomous robots in the field of robotics and mechatronics. One of the most important fields in that area is path planning for robots. Mobile robot path planning is a problem to find an optimal feasible path between starting point and target point with some obstacles in the path. Mobile robot path planning can be run at static or dynamic environment. In static environment the optimal path should determine before starting of the path planning algorithm. But in dynamic environment re-planning and updating are expected in a short time (such as every iteration). The significant problem in mobile robot path planning is that the algorithm can reach to local minimum instead of global minimum in finding optimal path, so choosing the optimizing algorithm affect as a vital factor in path planning [1-2]. Adem Tuncer proposed new mutation operator for the genetic algorithm and applied it to the path planning problem with mobile robots in dynamic environments. He showed path planning for a mobile robot can find a feasible path from a starting node to a target node in an environment with obstacles. GA is used to produce an optimal path by using the advantage of improved mutation operators to optimize the path planning [3]. Hong Qu presents a co-evolutionary improved genetic algorithm in path planning of multiple mobile robots globally, which operate a co-evolutionary method for optimization of the paths. He presents an effective and accurate fitness function, improves genetic operators and proposes and genetic modification operator in his paper. Furthermore, the proposed improved GA is compared with conventional GA [4]. J. Tu compared traditional search algorithms and optimization methods, such as calculus-based and numerical strategies with the evolutionary algorithms that are robust, globally and more useful when there is little or no prior knowledge about the problem to be solved [5]. One method that has more speed to the others in path planning is the GA method because of its capacity to explore the solution space that has a practical implementation on the FPGA [6]. Chuangling proposed a path planning method for mobile robots based on an adaptive genetic algorithm [7]. Kala presented a co-evolutionary genetic programming method to solve multi-robot path planning that includes different source and goal for each robot [8]. Other intelligent planning methods are used in a path planning environment with obstacles such as fuzzy logic [9] and neural networks [10]. Shirong Liu proposed a method with ant colony optimization algorithms in order to have distributed local navigation for multi-robot systems [11].

In this study, we use evolutionary algorithms for optimizing mobile robot path planning with dynamic environment. This approach could not have collided to obstacle in location of selected points and also the selected paths from starting to target point. The path planning should be optimized in the distance, smoothness and safety for proposed path globally in a grid-based environment. The optimized path planning algorithms are compared to each other in object function and time in some paths. The mentioned evolutionary methods are genetic algorithm, pattern search algorithm and particle swarm optimization.

## II. PROBLEM DESCRIPTION

For solving the mobile robot path planning, first we should define an environment that robot moves on it and then formulate the objective function based on the environment, robot characteristics and obstacles.

## A. Environment representation

In this paper, we use grid-based environment to represent the environment with feasible paths and infeasible paths. The grid-based environment includes obstacles and allowed locations in numbered grid. Our goal is to find locations on a grid-based environment in order that the robot can change its direction on those places only and minimize the object function between the starting point and target point. The important point that should be considered in path planning is neither selected position can choose on obstacle locations nor the selected path cross obstacle locations. In this paper grid-based environment includes 100 numbered grids with equality in width and length. Fig.1 illustrates the grid-based environment and shows a feasible path from the start point to target point. Also, we should say the obstacles can move on a grid-based environment this is the reason for dynamic environment path planning.

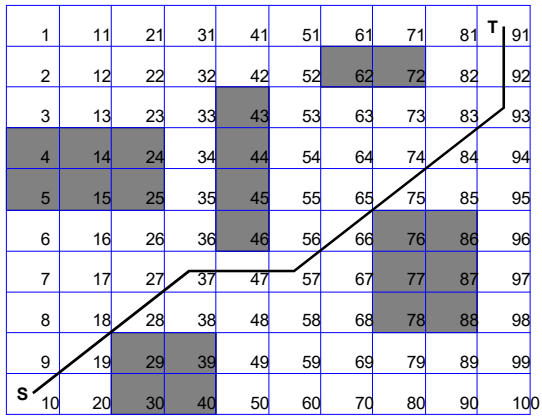

Fig. 1. Grid-based environment

## B. Objective function

The final goal in path planning problem is to find an optimal feasible path from starting point to target point. Optimality can interpret in various definitions like minimum fuel consumption, minimum energy consumption, minimum time spending, but here we use a combination of path distance, smoothness and safety for optimizing of the paths [4].

The objective function can compute in two manners as follows equations:

First for feasible paths:

$$f_{feas} = \alpha \sum_{i=1}^{n-1} d(P_i, P_{i+1}) + \beta \sum_{i=1}^{n-1} |\theta_i| + \gamma \ f_{semi} \quad (1)$$

$$f_{semi} = \begin{cases} C, & \text{if } \exists L_i, |L_i| < k, for \ i = 1, \dots, n-1 \\ 0, & \text{otherwise} \end{cases} \quad (2)$$

Second for infeasible paths:

$$f_{infeas} = f_{feas} + penalty \quad (3)$$

Where $f_{feas}$ is the objective function (OF) of the feasible paths, $f_{semi}$ is a term that we add it to the OF to avoid approaching to the certain vicinity of the obstacles (as the mobile robot has a certain width and can't pass from the edge of the obstacles). $f_{infeas}$ is the objective function of the infeasible paths. The feasible objective function is computed by the sum of distances between robot positions (points) in the path planning, sum of the angles that robot has in each point for smooth moving and an additional term for avoidance to the vicinity of the obstacle as an safety term. $P_i$ ($x_i$, $y_i$) is the current position of the robot and $P_{i+1}$ ($x_{i+1}$, $y_{i+1}$) is the next position of the robot. $\alpha$, $\beta$ and $\gamma$ are constant weights for adjusting the importance of their factors. We usually choose $\gamma$ higher than $\alpha$ and $\beta$ to avoid approaching to the obstacles as an important practical issue. $L_i$ is the minimum straight distance from $i^{th}$ path lines to the obstacles, C is chosen as a constant in the whole paths and k is selected as a parameter of safety margin from the obstacles. Penalty is added for obstacle avoidance and is chosen a large constant value in order to select feasible paths with the most chance (*penalty=1000*), n is the number of points for path planning. Also $d$ is Euclidean distance as follows:

$$d(P_i, P_{i+1}) = \sqrt{(x_{i+1} - x_i)^2 + (y_{i+1} - y_i)^2} \quad (4)$$

Where $x_i$ and $y_i$ is current horizontal and vertical positions of the robot and $x_{i+1}$ and $y_{i+1}$ is next horizontal and vertical positions of the robot. $L_i$ is the shortest distance between $i^{th}$ line and the obstacles. The direction of the robot is defined as below that is computed for each point in order to calculate the fitness function, $\theta_i$ is an angle in radian:

$$\theta_i = tan^{-1} \frac{(y_{i+1} - y_i)}{(x_{i+1} - x_i)} \quad (5)$$

## III. EVOLUTIONARY METHODS

In this section we introduce the mentioned evolutionary algorithms. Indeed, these are derivative free algorithms that are suitable to solve a variety of optimization problems that are not standard. We should use an effective tool for solving optimization problems in the absence of any information from the gradient of the objective function. Genetic algorithm, pattern search algorithm and particle swarm optimization are described as follows.

### A. PATTERN SEARCH ALGORITHM

Pattern search algorithm is a direct method for optimization of the objective function that is not differentiable, continuous and stochastic inevitably. Pattern search algorithm in despite of other optimization methods that are gradient base algorithms, it searches a set of points around the current point. In fact, it searches for the point that the objective function is lower than the others at the current point. In despite of other heuristic algorithms such as GA, it has a flexible operator to increase and adapt the global search. The iterative algorithm starts with the creation of a set of points that are called mesh near the current point. The mesh is created by adding the current point to multiple of vectors that are called pattern. Then, if a point in the mesh minimizes the object function more than others, the point at the first step of the pattern search algorithm becomes the current point at a subsequent level. The current point can be as an initial point or it can be supposed as a result of the previous step. Fig. 2 shows the formation of the Initial point, pattern and mesh points at 1$^{st}$ step of the pattern search algorithm. The algorithm is explained completely as below [12-15]:

1. The pattern search algorithm begins with initial point $x_0$. Initial point is updated every iteration.
2. Pattern vectors are created as [0 1], [1 0], [0 -1], [-1 0]. We have a scalar = 1 (iteration number) that is called mesh size.
3. Adding multiple of pattern vectors to the initial point, these new vectors are called mesh points. The multiple is mesh size here.
4. Computing the objective function at the mesh points.
5. Polling mesh points (by computing the objective function

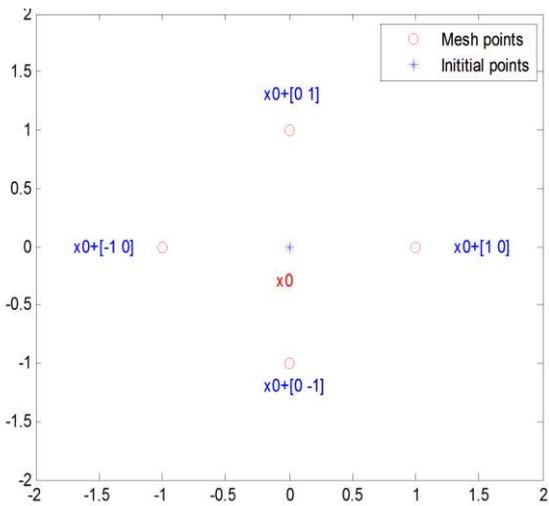

Fig. 2. Initial point, pattern and mesh points

at the mesh point and find one that it has smaller objective function value than the objective function value at initial point). If there is such point, successful poll happened. Depend on a successful poll or unsuccessful poll go to the phase 6 or 7.
6. After the successful poll, if the termination rule is not reached, the algorithm steps go to the next iteration and mesh size = mesh size × 2 and go to phase 3 and update the initial point as a mesh point of minimum objective function, otherwise go to phase 8.
7. After unsuccessful poll and it means none of the mesh point has a smaller objective function value than the objective function value at the initial point. If the termination rule is not reached, the algorithm steps go to the next iteration and mesh size = mesh size × 0.5 and go to phase 3 and update the initial point as previous mesh point, otherwise go to phase 8.
8. Termination rule is reached and the global variables are found based on minimizing the objective function. The patterns search algorithm terminates if the number of iterations reaches to max iteration or mesh size is less than mesh limit or the changing between the point that find at one successful poll and next successful poll is less than X limit or changing the objective function between one successful poll and next successful poll is less than the objective function limit [13].

A. GENETIC ALGORITHM (GA)

GA is an iterative random global search method that it can find the solution of the problems with inspired processes from the natural evolution. We can apply the GA to many difficult optimization problems perfectly. It shows its preferences over traditional optimization algorithms. If the problem has multiple optimum solutions, it can find the global solution. GA presents a large number of candidate solutions to a problem that are called as population at each iteration. Every solution is shown by a string, which is named as chromosome that is usually coded as a binary string. The chromosome has been decoded to real numbers and after that the objective function of each chromosome is computed as performance function. The most important problems in GA are finding the correct function that is called as objective function or fitness function and the genetic coding in order to determine the problem specification correctly. The initial population in GA is made randomly and they are used in the next iteration by genetic operators: selection method, crossover and mutation operator and termination conditions. The selection method chooses the chromosomes that have a higher fitness function. They appear with higher probability in the next iteration or generation. Crossover operator is executed between two selected chromosomes that are called as parents in order to exchange parts of their chromosomes. This operator starts from a random crossover point and tends to make possible the evolutionary process to move to the vast regions of the search solutions. Mutation operator is used in problem solutions to avoid local optimum, it changes zero to one or one to zero on the chromosomes in random places. Here, termination rule is the limited number of iterations that the algorithm is repeated [16].

B. PARTICLE SWARM OPTIMIZATION

Particle swarm optimization is a heuristic search method that first proposed by Kennedy and Eberhart in 1995. This evolutionary computational method is used for dealing with the optimization of continuous and discontinuous objective function. The PSO algorithm is inspired from the biological and sociological behavior of some animals such as groups of fishes and flocks of birds that are searching for their food. PSO is similar to GA in some part. It is also population-based search method that each solution is shown as a particle in a population, which is called swarm. Particles can change their positions in a multi-dimensional search space in order to reach their equilibrium and optimal points until iteration limitation is occurring. Swarm of N particles is initialized, which each particle is allocated a randomly. $x$ is symbolized as a particle's position and $v$ denote the particle's velocity over multi-dimensional solution space. Each individual $x$ in the swarms is ranked by using an objective function that illustrates how well it solves the problem. The index of the best previous position of a particle is *Pbest*. The index of the best particle among all particles in the swarm until this iteration is *Gbest*. Each particle records its own personal best position by index of *Pbest* and it is compared to the best positions that are found by all particles in the swarm by index of *Gbest*. Velocity and position of a particle are updated in each iteration as the follows [16,17]:

$$v_i(t) = w1.v_i(t-1) + c1\left(x_{Pbest_i} - x_i(t)\right) + c2\left(x_{Gbest} - x_i(t)\right) \quad (6)$$

$$x_i(t) = x_i(t-1) + v_i(t) \quad (7)$$

$w1$, $c1$ and $c2$ are as constant weights.

Short illustrations of the iterative evolutionary algorithms are shown as a flowchart in Fig. 3. GA, PS and PSO are the abbreviation of the genetic algorithm, pattern search algorithm and particle swarm optimization. The evolutionary algorithm will repeat the mentioned algorithm until it finds the optimal result that it minimizes the objective function globally.

IV. SIMULATION RESULTS AND DISCUSSION

The grid-based environments include 5 obstacles (shaded area) and 10 × 10 grids. The proposed methods are coded in MATLAB in a PC with the specifications; CPU: Intel-Core 2 Duo-2.4GHz, RAM: 3GB in 32-bit operating system. The constants and weights of the equations are chosen arbitrary. Two different examples of path planning in a grid - based environment are compared in order to show the success of the proposed algorithm. First, we demonstrate and compare the initial environment path planning with GA, PS and PSO in which environment is not changed in obstacles and then in a modified environment. Fig.4, Fig.5 and Fig.6 demonstrate a path planning for mobile robot using GA, PS and PSO algorithms in the initial environment respectively. It is obvious that each algorithm chooses a different path for reaching to target point and paths do not cross obstacles. Since path planning of three algorithms has different, we should examine each path with a special obstacle moving as a dynamic environment. In order to have dynamic environment, we move the obstacle at location 62 and 72 to location 58 and 68 as shown in Fig.7. Thus, the mobile robot should change its default path planning and update it in order to avoid collision. Fig.7, Fig.8 and Fig.9 show path planning for mobile robot using GA, PS and PSO algorithms in the modified environment respectively. Fig.10 demonstrates the Path planning using evolutionary algorithms in z-type environment. It is obvious that we can compare the start-target distance, smoothness and safety in the three evolutionary algorithms as shown in Fig.10. Table 1 shows the comparison between the evolutionary algorithms in time and the objective function. It is obviously demonstrated the PSO has better results in both initial and modified environments in the objective function, although the PS has the minimum time for the initial and modified environment. GA has the worse results compared to PS and PSO. Table 2 illustrates the comparison of the mentioned algorithms in z-type environment. As it shows the PSO has the best objective function and the PS has the lowest in time.

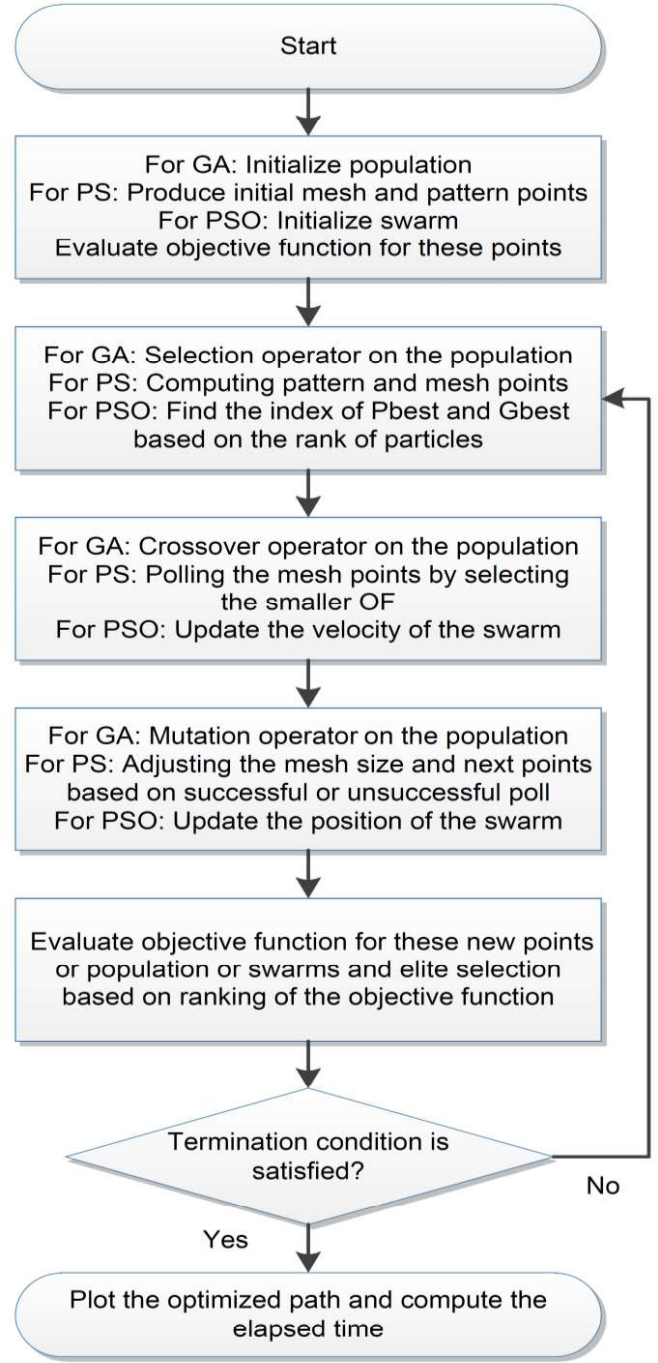

Fig. 3. Flowchart of iterative evolutionary algorithms.

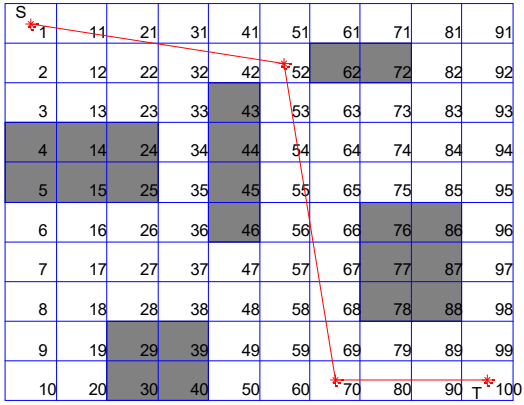

Fig. 4 Path planning using GA in the initial environment

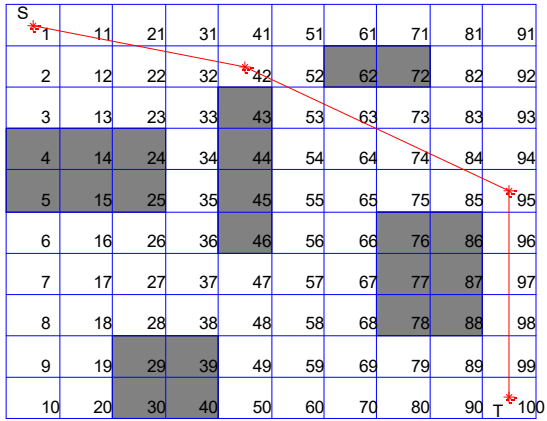

Fig. 5 Path planning using PS in the initial environment

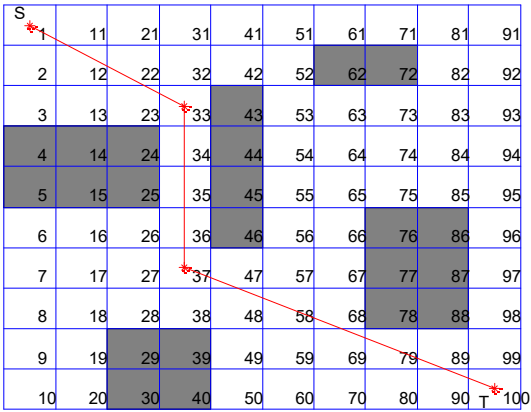

Fig. 6 Path planning using PSO in the initial environment

## V. CONCLUSION

In this research, we introduce evolutionary algorithms for optimization of path planning of mobile robots in a dynamic grid-based environment. In order to show the success of the method, three examples in the initial and modified environment are compared with each other. Both initial and modified environments PSO has better results in the objective function than the other algorithms. This result also is repeated the in the z-type environment. The results show PS is more useful in the reduction of the optimization time in the path planning problem. Therefore, this study shows the PSO is better in accuracy and minimization of the objective function, while the PS has more speed to converge. So PSO and PS can be effective methods for optimization of the path planning mobile robot in a suitable situation.

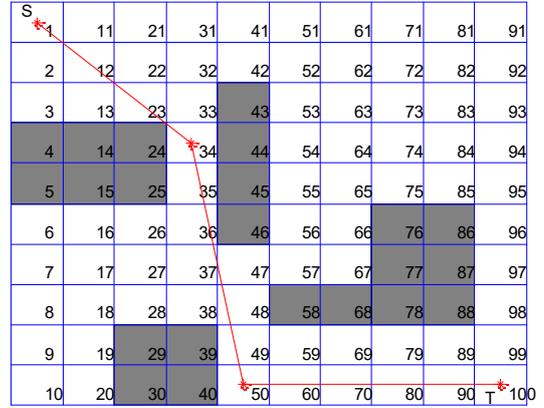

Fig. 7 Path planning using GA in the modified environment

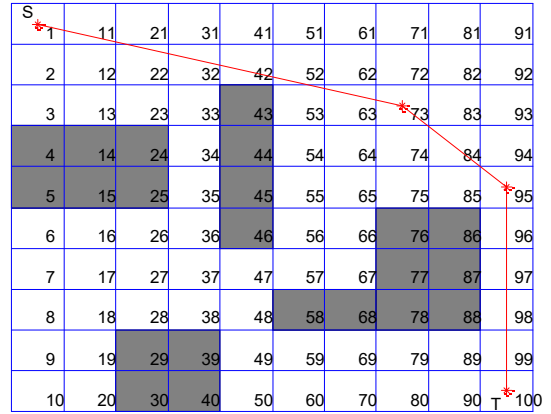

Fig. 8 Path planning using PS in the modified environment

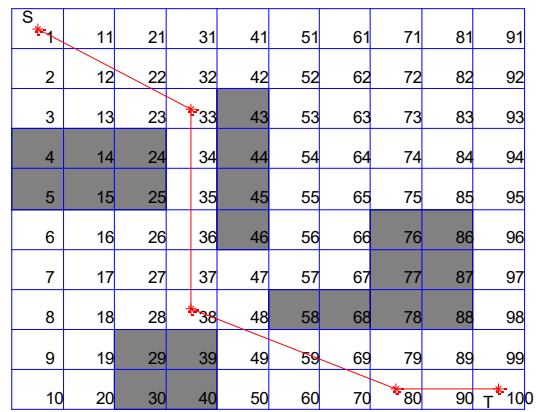

Fig. 9 Path planning using PSO in the modified environment

Table 1. Comparison results in evolutionary methods for mobile robot path planning in the initial and modified environment

| Algorithms | Objective function at initial environment | Objective function at modified environment | Time for initial environment (sec) | Time for modified environment (sec) |
|---|---|---|---|---|
| Genetic Algorithm | 17.8051 | 18.4410 | 0.9310 | 1.1180 |
| Pattern Search Algorithm | 17.3103 | 17.7430 | 0.6800 | 0.8310 |
| Particle Swarm Optimization | 16.9362 | 17.7001 | 2.2300 | 2.5590 |

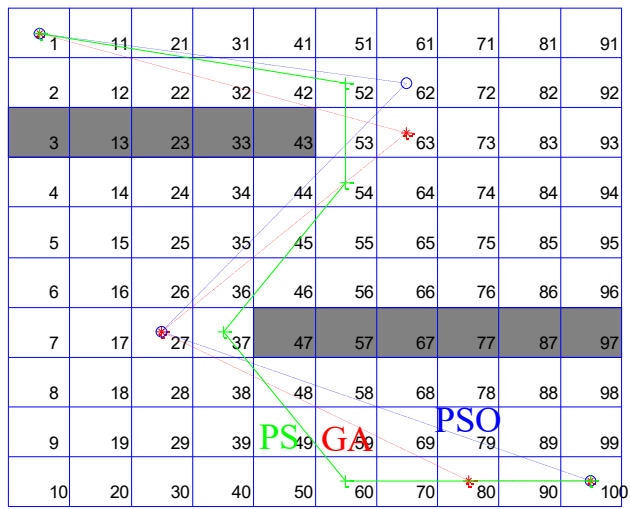

Fig. 10 Path planning using evolutionary algorithms in z-type environment

Table 2. Comparison results in evolutionary methods for z-type environment

| Algorithms | Objective function | Time (sec) |
|---|---|---|
| Genetic Algorithm | 22.4599 | 0.9310 |
| Pattern Search Algorithm | 22.0439 | 0.7879 |
| Particle Swarm Optimization | 21.9185 | 2.5920 |